\documentclass{article}

\usepackage{times}
\usepackage{graphicx} % more modern
\usepackage{subfigure}

\usepackage{natbib}

\usepackage{algorithm}
\usepackage{algorithmic}

\usepackage{hyperref}

\usepackage[accepted]{icml2015} 

\usepackage[titletoc,title]{appendix}

\usepackage{amsmath}
\usepackage{amssymb}
\usepackage{amsfonts}
\usepackage{dsfont}
\usepackage{xspace}
\usepackage{xr}

\usepackage{multirow}
\usepackage{siunitx}
\usepackage{tikz}
\usetikzlibrary{positioning}
\usetikzlibrary{calc}

\usepackage{pgfplots}
\pgfplotsset{compat=newest} 
\pgfplotsset{plot coordinates/math parser=false}

\newlength\figureheight
\newlength\figurewidth

\newcommand{\fig}[1]{Figure~\ref{fig:#1}}
\newcommand{\sect}[1]{Section~\ref{sect:#1}}
\newcommand{\tab}[1]{Table~\ref{tab:#1}}
\newcommand{\eq}[1]{(\ref{eq:#1})}

\makeatletter
\newcommand{\todo}[1][]{\@latex@warning{TODO #1}\fbox{TODO\dots}}
\makeatother

\icmltitlerunning{Unsupervised Domain Adaptation by Backpropagation}

\begin{document} 

\twocolumn[
\icmltitle{Unsupervised Domain Adaptation by Backpropagation}

% unless you've provided the [accepted] option to the icml2015
\icmlauthor{Yaroslav Ganin}{ganin@skoltech.ru}
\icmladdress{\vskip -0.3cm}
\icmlauthor{Victor Lempitsky}{lempitsky@skoltech.ru}
\icmladdress{Skolkovo Institute of Science and Technology (Skoltech)}

\icmlkeywords{Gradient Reversal, Unsupervised Domain Adaptation, Deep Learning}

\vskip 0.3in
]

\begin{abstract} 
Top-performing deep architectures are trained on massive amounts of labeled data. In the absence of labeled data for a certain task, domain adaptation often provides an attractive option given that labeled data of similar nature but from a different domain (e.g.\ synthetic images) are available. Here, we propose a new approach to domain adaptation in deep architectures that can be trained on large amount of labeled data from the source domain and large amount of unlabeled data from the target domain (no labeled target-domain data is necessary). 

As the training progresses, the approach promotes the emergence of ``deep'' features that are (i) discriminative for the main learning task on the source domain and (ii) invariant with respect to the shift between the domains. We show that this adaptation behaviour can be achieved in almost any feed-forward model by augmenting it with few standard layers and a simple new {\em gradient reversal} layer. The resulting augmented architecture can be trained using standard backpropagation.

Overall, the approach can be implemented with little effort using any of the deep-learning packages. The method performs very well in a series of image classification experiments, achieving adaptation effect in the presence of big domain shifts and outperforming previous state-of-the-art on Office datasets.\vspace{-1.5mm}
\end{abstract} 

\section{Introduction}

Deep feed-forward architectures have brought impressive advances to the state-of-the-art across a wide variety of machine-learning tasks and applications. At the moment, however, these leaps in performance come only when a large amount of labeled training data is available. At the same time, for problems lacking labeled data, it may be still possible to obtain training sets that are big enough for training large-scale deep models, but that suffer from the {\em shift} in data distribution from the actual data encountered at ``test time''. One particularly important example is synthetic or semi-synthetic training data, which may come in abundance and be fully labeled, but which inevitably have a distribution that is different from real data~\cite{Liebelt10,Stark10,Vazquez14,Sun14}.

Learning a discriminative classifier or other predictor in the presence of a {\em shift} between training and test distributions is known as {\em domain adaptation} (DA). A number of approaches to domain adaptation has been suggested in the context of {\em shallow} learning, e.g.\ in the situation when data representation/features are given and fixed. The proposed approaches then build the mappings between the {\em source} (training-time) and the {\em target} (test-time) domains, so that the classifier learned for the source domain can also be applied to the target domain, when composed with the learned mapping between domains. The appeal of the domain adaptation approaches is the ability to learn a mapping between domains in the situation when the target domain data are either fully unlabeled ({\em unsupervised domain annotation}) or have few labeled samples ({\em semi-supervised domain adaptation}). Below, we focus on the harder unsupervised case, although the proposed approach can be generalized to the semi-supervised case rather straightforwardly.

Unlike most previous papers on domain adaptation that worked with fixed feature representations, we focus on combining domain adaptation and deep feature learning within one training process ({\em deep domain adaptation}). Our goal is to embed domain adaptation into the process of learning representation, so that the final classification decisions are made based on features that are both discriminative and invariant to the change of domains, i.e.\ have the same or very similar distributions in the source and the target domains. In this way, the obtained feed-forward network can be applicable to the target domain without being hindered by the shift between the two domains.

We thus focus on learning features that combine (i) discriminativeness and (ii) domain-invariance. This is achieved by jointly optimizing the underlying features as well as two discriminative classifiers operating on these features: (i) the \emph{label predictor} that predicts class labels and is used both during training and at test time and (ii) the \emph{domain classifier} that discriminates between the source and the target domains during training. While the parameters of the classifiers are optimized in order to minimize their error on the training set, the parameters of the underlying deep feature mapping are optimized in order to {\em minimize} the loss of the label classifier and to {\em maximize} the loss of the domain classifier. The latter encourages domain-invariant features to emerge in the course of the optimization.

\begin{figure*}

\centering
\includegraphics[width=0.8\textwidth]{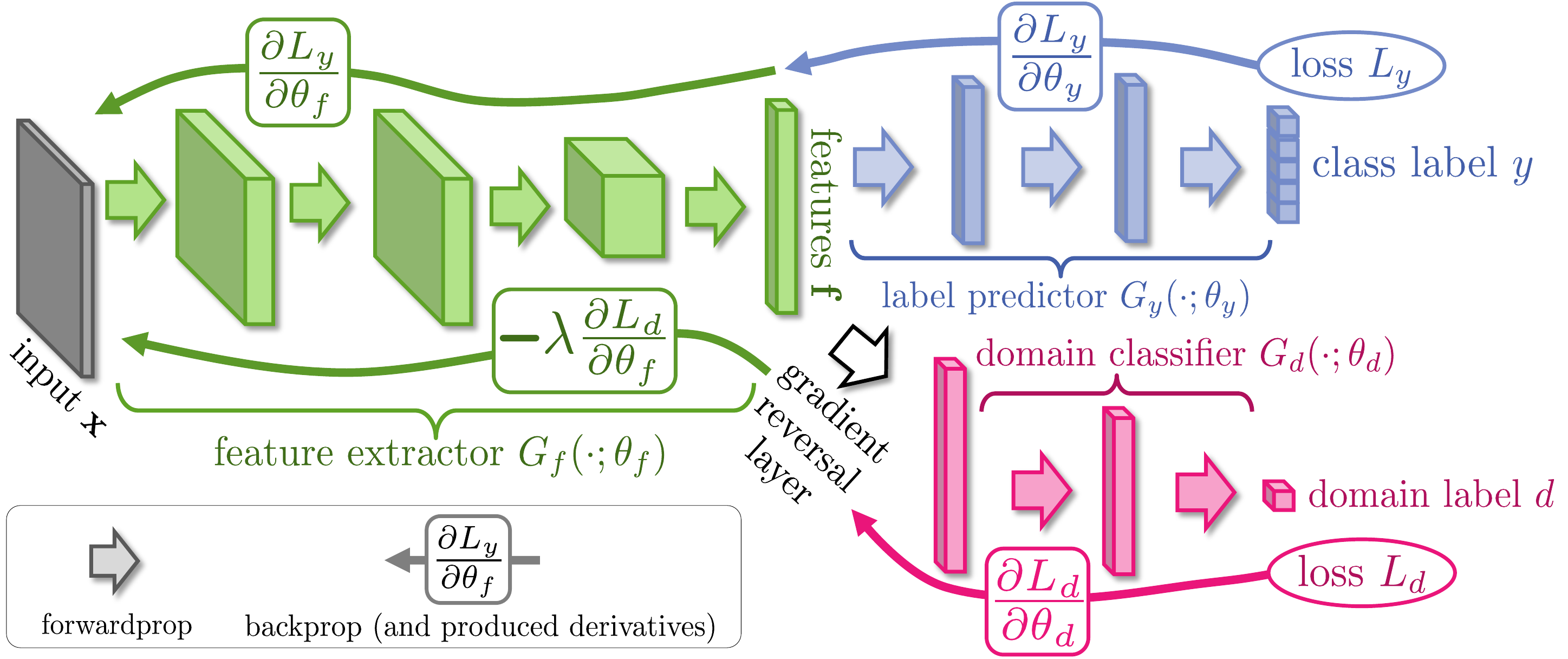}

\caption{The {\bf proposed architecture} includes a deep {\em feature extractor} (green) and a deep {\em label predictor} (blue), which together form a standard feed-forward architecture. Unsupervised domain adaptation is achieved by adding a {\em domain classifier} (red) connected to the feature extractor via a {\em gradient reversal layer} that multiplies the gradient by a certain negative constant during the backpropagation-based training. Otherwise, the training proceeds in a standard way and minimizes the label prediction loss (for source examples) and the domain classification loss (for all samples). Gradient reversal ensures that the feature distributions over the two domains are made similar (as indistinguishable as possible for the domain classifier), thus resulting in the domain-invariant features.\vspace{-2mm}}
\label{fig:arch}
\end{figure*}

Crucially, we show that all three training processes can be embedded into an appropriately composed deep feed-forward network (\fig{arch}) that uses standard layers and loss functions, and can be trained using standard backpropagation algorithms based on stochastic gradient descent or its modifications (e.g.\ SGD with momentum). Our approach is generic as it can be used to add domain adaptation to any existing feed-forward architecture that is trainable by backpropagation. In practice, the only non-standard component of the proposed architecture is a rather trivial {\em gradient reversal} layer that leaves the input unchanged during forward propagation and reverses the gradient by multiplying it by a negative scalar during the backpropagation.  

%As will be seen, such  layer allows the optimization process to mimimize the ability of the domain classifier to discriminate between the domains and ensures the domain-invariance of the deep features.

Below, we detail the proposed approach to domain adaptation in deep architectures, and present results on traditional deep learning image datasets (such as MNIST~\cite{LeCun98} and SVHN~\cite{Netzer11}) as well as on {\sc Office} benchmarks \cite{Saenko10}, where the proposed method considerably improves over previous state-of-the-art accuracy.

\section{Related work}
\label{sec:related}

A large number of domain adaptation methods have been proposed over the recent years, and here we focus on the most related ones. Multiple methods perform unsupervised domain adaptation by matching the feature distributions in the source and the target domains. Some approaches perform this by reweighing or selecting samples from the source domain \cite{Borgwardt06,Huang06,Gong13}, while others seek an explicit feature space transformation that would map source distribution into the target ones \cite{Pan11,Gopalan11,Baktashmotlagh13}. An important aspect of the distribution matching approach is the way the (dis)similarity between distributions is measured.  Here, one popular choice is matching the distribution means in the kernel-reproducing Hilbert space \cite{Borgwardt06,Huang06}, whereas \cite{Gong12,Fernando13} map the principal axes associated with each of the distributions. Our approach also attempts to match feature space distributions, however this is accomplished by modifying the feature representation itself rather than by reweighing or geometric transformation. Also, our method uses (implicitly) a rather different way to measure the disparity between distributions based on their separability by a deep discriminatively-trained classifier.

Several approaches perform gradual transition from the source to the target domain \cite{Gopalan11,Gong12} by a gradual change of the training distribution. Among these methods, \cite{Chopra13} does this in a ``deep'' way by the layerwise training of a sequence of deep autoencoders, while gradually replacing source-domain samples with target-domain samples. This improves over a similar approach of \cite{Glorot11} that simply trains a single deep autoencoder for both domains. In both approaches, the actual classifier/predictor is learned in a separate step using the feature representation learned by autoencoder(s). In contrast to \cite{Glorot11,Chopra13}, our approach performs feature learning, domain adaptation and classifier learning jointly, in a unified architecture, and using a single learning algorithm (backpropagation). We therefore argue that our approach is simpler (both conceptually and in terms of its implementation). Our method also achieves considerably better results on the popular {\sc Office} benchmark.

While the above approaches perform unsupervised domain adaptation, there are approaches that perform {\em supervised} domain adaptation by exploiting labeled data from the target domain. In the context of deep feed-forward architectures, such data can be used to ``fine-tune'' the network trained on the source domain~\cite{Zeiler13,Oquab14,Babenko14}. Our approach does not require labeled target-domain data. At the same time, it can easily incorporate such data when it is available.

An idea related to ours is described in \cite{Goodfellow14}. While their goal is quite different (building generative deep networks that can synthesize samples), the way they measure and minimize the discrepancy between the distribution of the training data and the distribution of the synthesized data is very similar to the way our architecture measures and minimizes the discrepancy between feature distributions for the two domains.

Finally, a recent and concurrent report by \cite{Tzeng14} also focuses on domain adaptation in feed-forward networks. Their set of techniques measures and minimizes the distance of the data means across domains. This approach may be regarded as a ``first-order'' approximation to our approach, which seeks a tighter alignment between distributions.
\section{Deep Domain Adaptation}

\def\x{{\mathbf x}}
\def\f{{\mathbf f}}

\def\S{{\cal S}}
\def\T{{\cal T}}

\def\R{{\mathds R}}

\def\tf{{\theta_f}}
\def\td{{\theta_d}}
\def\ty{{\theta_y}}
\def\htf{{\hat\theta_f}}
\def\htd{{\hat\theta_d}}
\def\hty{{\hat\theta_y}}

\subsection{The model}

We now detail the proposed model for the domain adaptation. We assume that the model works with input samples $\x \in X$, where $X$ is some input space and certain labels (output) $y$ from the label space $Y$. Below, we assume classification problems where $Y$ is a finite set ($Y=\{1,2,\dots L\}$), however our approach is generic and can handle any output label space that other deep feed-forward models can handle. We further assume that there exist two distributions $\S(x,y)$ and $\T(x,y)$ on $X\otimes Y$, which will be referred to as the source distribution and the target distribution (or the source domain and the target domain). Both distributions are assumed complex and unknown, and furthermore similar but different (in other words, $\S$ is ``shifted'' from $\T$ by some {\em domain shift}). 

Our ultimate goal is to be able to predict labels $y$ given the input $\x$ for the target distribution. At training time, we have an access to a large set of training samples $\{\x_1,\x_2,\dots,\x_N\}$ from both the source and the target domains distributed according to the marginal distributions $\S(\x)$ and $\T(\x)$. We denote with $d_i$ the binary variable ({\em domain label}) for the $i$-th example, which indicates whether $x_i$ come from the source distribution ($\x_i{\sim}\S(\x)$ if $d_i{=}0$) or from the target distribution ($\x_i{\sim}\T(\x)$ if $d_i{=}1$). For the examples from the source distribution ($d_i{=}0$) the corresponding labels $y_i \in Y$ are known at training time. For the examples from the target domains, we do not know the labels at training time, and we want to predict such labels at test time.

We now define a deep feed-forward architecture that for each input $\x$ predicts its label $y \in Y$ {\em and} its domain label $d \in \{0,1\}$. We decompose such mapping into three parts. We assume that the input $\x$ is first mapped by a mapping $G_f$ (a {\em feature extractor}) to a $D$-dimensional feature vector $\f \in \R^D$. The feature mapping may also include several feed-forward layers and we denote the vector of parameters of all layers in this mapping as $\tf$, i.e.\ $\f = G_f(\x;\tf)$. Then, the feature vector $\f$ is mapped by a mapping $G_y$ ({\em label predictor}) to the label $y$, and we denote the parameters of this mapping with $\ty$. Finally, the same feature vector $\f$ is mapped to the domain label $d$ by a mapping $G_d$ ({\em domain classifier}) with the parameters $\td$ (\fig{arch}).

During the learning stage, we aim to minimize the label prediction loss on the annotated part (i.e.\ the source part) of the training set, and the parameters of both the feature extractor and the label predictor are thus optimized in order to minimize the empirical loss for the source domain samples. This ensures the discriminativeness of the features $\f$ and the overall good prediction performance of the combination of the feature extractor and the label predictor on the source domain.

At the same time, we want to make the features $\f$ domain-invariant. That is, we want to make the distributions $S(\f) = \{G_f(\x; \tf)\,|\, \x{\sim}S(\x)\}$ and $T(\f) = \{G_f(\x; \tf)\,|\, \x{\sim}T(\x)\}$ to be similar. Under the {\em covariate shift} assumption, this would make the label prediction accuracy on the target domain to be the same as on the source domain~\cite{Shimodaira00}. Measuring the dissimilarity of the distributions $S(\f)$ and $T(\f)$ is however non-trivial, given that $\f$ is high-dimensional, and that the distributions themselves are constantly changing as learning progresses. One way to estimate the dissimilarity is to look at the loss of the domain classifier $G_d$, provided that the parameters $\td$ of the domain classifier have been trained to discriminate between the two feature distributions in an optimal way. 

This observation leads to our idea. At training time, in order to obtain domain-invariant features, we seek the parameters $\tf$ of the feature mapping that {\em maximize} the loss of the domain classifier (by making the two feature distributions as similar as possible), while simultaneously seeking the parameters $\td$ of the domain classifier that {\em minimize} the loss of the domain classifier. In addition, we seek to minimize the loss of the label predictor. 

More formally, we consider the functional:
\begin{gather} 
E(\tf,\ty,\td) =  \sum_{\substack{i=1..N\\d_i = 0}} L_y\left( \strut G_y(G_f(\x_i;\tf);\ty), y_i\right) -\nonumber\\ \lambda \sum_{i=1..N} L_d \left( \strut G_d(G_f(\x_i;\tf);\td), y_i\right) = \nonumber\\= \sum_{\substack{i=1..N\\d_i = 0}} L^i_y( \tf, \ty) - \lambda \sum_{i=1..N} L^i_d( \tf, \td) 
\label{eq:func}
\end{gather}
Here, $L_y(\cdot,\cdot)$ is the loss for label prediction (e.g.\ multinomial), $L_d(\cdot,\cdot)$ is the loss for the domain classification (e.g.\ logistic), while $L^i_y$ and $L^i_d$ denote the corresponding loss functions evaluated at the $i$-th training example. 

Based on our idea, we are seeking the parameters $\htf,\hty,\htd$ that deliver a saddle point of the functional \eq{func}:

\begin{gather}
(\htf,\hty) = \arg\min_{\tf,\ty} E(\tf,\ty,\htd)\,\label{eq:opt1}\\\label{eq:opt2}
\htd = \arg\max_\td E(\htf,\hty, \td)\,.
\end{gather}

At the saddle point, the parameters $\td$ of the domain classifier $\td$ minimize the domain classification loss (since it enters into \eq{func} with the minus sign) while the parameters $\ty$ of the label predictor minimize the label prediction loss. The feature mapping parameters $\tf$ {\em minimize} the label prediction loss (i.e.\ the features are discriminative), while {\em maximizing} the domain classification loss (i.e.\ the features are domain-invariant). The parameter $\lambda$ controls the trade-off between the two objectives that shape the features during learning.

Below, we demonstrate that standard stochastic gradient solvers (SGD) can be adapted for the search of the saddle point \eq{opt1}-\eq{opt2}.

\subsection{Optimization with backpropagation}

A saddle point \eq{opt1}-\eq{opt2} can be found as a stationary point of the following stochastic updates:

\begin{align}
\tf \quad &\longleftarrow \quad \tf \;-\; \mu \left(\frac{\partial L^i_y}{\partial \tf}-\lambda\frac{\partial L^i_d}{\partial \tf} \right) \label{eq:upd1}\\
\ty \quad &\longleftarrow \qquad \ty \;-\; \mu \frac{\partial L^i_y}{\partial \ty}\label{eq:upd2}\\
\td \quad &\longleftarrow \qquad \td \;-\; \mu \frac{\partial L^i_d}{\partial \td} \label{eq:upd3}
\end{align}
where $\mu$ is the learning rate (which can vary over time).

The updates \eq{upd1}-\eq{upd3} are very similar to stochastic gradient descent (SGD) updates for a feed-forward deep model that comprises feature extractor fed into the label predictor and into the domain classifier. The difference is the $-\lambda$ factor in \eq{upd1} (the difference is important, as without such factor, stochastic gradient descent would try to make features dissimilar across domains in order to minimize the domain classification loss). Although direct implementation of \eq{upd1}-\eq{upd3} as SGD is not possible, it is highly desirable to reduce the updates \eq{upd1}-\eq{upd3} to some form of SGD, since SGD (and its variants) is the main learning algorithm implemented in most packages for deep learning. 

Fortunately, such reduction can be accomplished by introducing a special {\bf gradient reversal layer} (GRL) defined as follows. The gradient reversal layer has no parameters associated with it (apart from the meta-parameter $\lambda$, which is not updated by backpropagation). During the forward propagation, GRL acts as an identity transform. During the backpropagation though, GRL takes the gradient from the subsequent level, multiplies it by $-\lambda$ and passes it to the preceding layer. Implementing such layer using existing object-oriented packages for deep learning is simple, as defining procedures for forwardprop (identity transform), backprop (multiplying by a constant), and parameter update (nothing) is trivial. 

The GRL as defined above is inserted between the feature extractor and the domain classifier, resulting in the architecture depicted in \fig{arch}. As the backpropagation process passes through the GRL, the partial derivatives of the loss that is downstream the GRL (i.e.\ $L_d$) w.r.t.\ the layer parameters that are upstream the GRL (i.e.\ $\tf$) get multiplied by $-\lambda$, i.e.\ $\frac{\partial L_d}{\partial \tf}$ is effectively replaced with $-\lambda\frac{\partial L_d}{\partial \tf}$. Therefore, running SGD in the resulting model implements the updates \eq{upd1}-\eq{upd3} and converges to a saddle point of \eq{func}.

Mathematically, we can formally treat the gradient reversal layer as a ``pseudo-function'' $R_\lambda(\x)$ defined by two (incompatible) equations describing its forward- and backpropagation behaviour:
\begin{align}
R_\lambda(\x) = \x\\
\frac{dR_\lambda}{d\x} = -\lambda \mathbf{I}
\end{align}
where $\mathbf{I}$ is an identity matrix.
We can then define the objective ``pseudo-function'' of $(\tf,\ty,\td)$ that is being optimized by the stochastic gradient descent within our method:
\begin{gather}
\tilde E(\tf,\ty,\td) =  \sum_{\substack{i=1..N\\d_i = 0}} L_y\left( \strut G_y(G_f(\x_i;\tf);\ty), y_i\right) +\nonumber\\ \sum_{i=1..N} L_d \left( \strut G_d(R_\lambda(G_f(\x_i;\tf));\td), y_i\right) \label{eq:pseudoobj}
\end{gather}

Running updates \eq{upd1}-\eq{upd3} can then be implemented as doing SGD for \eq{pseudoobj} and leads to the emergence of features that are domain-invariant and discriminative at the same time. After the learning, the label predictor $y(\x) = G_y(G_f(\x;\tf);\ty)$ can be used to predict labels for samples from the target domain (as well as from the source domain).

The simple learning procedure outlined above can be re-derived/generalized along the lines suggested in \cite{Goodfellow14} (see \ref{sect:appendix_alternative}).

\subsection{Relation to $ \mathcal{H} \Delta \mathcal{H} $-distance}
\label{sect:theory}

In this section we give a brief analysis of our method in terms of $ \mathcal{H} \Delta \mathcal{H} $-distance \cite{Ben10,Cortes11} which is widely used in the theory of non-conservative domain adaptation. Formally,
\begin{multline}
  d_{\mathcal{H} \Delta \mathcal{H}} (\mathcal{S}, \mathcal{T}) = 2 \sup_{h_1, h_2 \in \mathcal{H}} \left| P_{\mathbf{f} \sim \mathcal{S}} [h_1(\mathbf{f}) \neq h_2(\mathbf{f})] - \right. \\
  \left. - P_{\mathbf{f} \sim \mathcal{T}} [h_1(\mathbf{f}) \neq h_2(\mathbf{f})] \right|
\label{eq:hdh_dist}
\end{multline}
defines a discrepancy distance between two distributions $ \mathcal{S} $ and $ \mathcal{T} $ w.r.t. a hypothesis set $ \mathcal{H} $. Using this notion one can obtain a probabilistic bound \cite{Ben10} on the performance $ \varepsilon_\mathcal{T}(h) $ of some classifier $ h $ from $ \mathcal{T} $ evaluated on the target domain given its performance $ \varepsilon_\mathcal{S}(h) $ on the source domain:
\begin{equation}
  \varepsilon_\mathcal{T}(h) \leq \varepsilon_\mathcal{S}(h) + \frac{1}{2} d_{\mathcal{H} \Delta \mathcal{H}} (\mathcal{S}, \mathcal{T}) + C \, ,
\end{equation}
where $ \mathcal{S} $ and $ \mathcal{T} $ are source and target distributions respectively, and $ C $ does not depend on particular $ h $. 

Consider fixed $ \mathcal{S} $ and $ \mathcal{T} $ over the representation space produced by the feature extractor $ G_f $ and a family of label predictors $ \mathcal{H}_p $. We assume that the family of domain classifiers $ \mathcal{H}_d $ is rich enough to contain the symmetric difference hypothesis set of $ \mathcal{H}_p $:
\begin{equation}
  \mathcal{H}_p \Delta \mathcal{H}_p = \left\{ h \, | \, h = h_1 \oplus h_2 \, , \; h_1, h_2 \in \mathcal {H}_p \right\} \, .
\end{equation}
It is not an unrealistic assumption as we have a freedom to pick $ \mathcal{H}_d $ whichever we want. For example, we can set the architecture of the domain discriminator to be the layer-by-layer concatenation of two replicas of the label predictor followed by a two layer non-linear perceptron aimed to learn the \texttt{XOR}-function. Given the assumption holds, one can easily show that training the $ G_d $ is closely related to the estimation of $ d_{\mathcal{H}_p \Delta \mathcal{H}_p} (\mathcal{S}, \mathcal{T}) $. Indeed, 
\begin{equation}
\begin{split}
  &d_{\mathcal{H}_p \Delta \mathcal{H}_p}  (\mathcal{S}, \mathcal{T}) =\\ 
  & = 2 \sup_{h \in \mathcal{H}_p \Delta \mathcal{H}_p} \left| P_{\mathbf{f} \sim \mathcal{S}} [h(\mathbf{f}) = 1] - P_{\mathbf{f} \sim \mathcal{T}} [h(\mathbf{f}) = 1] \right| \leq \\
  & \leq 2 \sup_{h \in \mathcal{H}_d} \left| P_{\mathbf{f} \sim \mathcal{S}} [h(\mathbf{f}) = 1] - P_{\mathbf{f} \sim \mathcal{T}} [h(\mathbf{f}) = 1] \right| = \\
  & = 2 \sup_{h \in \mathcal{H}_d} \left| 1 - \alpha(h) \right| = 2 \sup_{h \in \mathcal{H}_d} \left[ \alpha(h) - 1 \right]
\end{split}
\end{equation}
where $ \alpha(h) = P_{\mathbf{f} \sim \mathcal{S}} [h(\mathbf{f}) = 0] + P_{\mathbf{f} \sim \mathcal{T}} [h(\mathbf{f}) = 1] $ is maximized by the optimal $ G_d $.

Thus, optimal discriminator gives the upper bound for $ d_{\mathcal{H}_p \Delta \mathcal{H}_p}  (\mathcal{S}, \mathcal{T}) $. At the same time, backpropagation of the reversed gradient changes the representation space so that $ \alpha(G_d) $ becomes smaller effectively reducing $ d_{\mathcal{H}_p \Delta \mathcal{H}_p}  (\mathcal{S}, \mathcal{T}) $ and leading to the better approximation of $ \varepsilon_\mathcal{T}(G_y) $ by $ \varepsilon_\mathcal{S}(G_y) $.

\section{Experiments}
\label{sect:experiments}

\begin{figure*}
  \centering
  \setlength{\tabcolsep}{0pt}
  \setlength\figurewidth{0.05\textwidth}
  \newcommand{\example}[1]{\raisebox{-.4\height}{\includegraphics[width=\figurewidth]{./figures/domains_examples/#1}}}
  \begin{sc}
  \begin{small}
  \begin{tabular}{r@{\hskip 0.5cm} ccc c@{\hskip 0.4cm} ccc c@{\hskip 0.4cm} ccc c@{\hskip 0.4cm} ccc}
    &
    \multicolumn{3}{c}{MNIST} & &
    \multicolumn{3}{c}{Syn Numbers} & &
    \multicolumn{3}{c}{SVHN} & &
    \multicolumn{3}{c}{Syn Signs}\\
    
    Source &
    \example{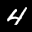} &
    \example{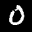} &
    \example{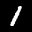} & &
    
    \example{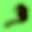} &
    \example{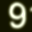} &
    \example{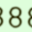} & &
    
    \example{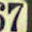} &
    \example{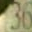} &
    \example{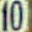} & &
    
    \example{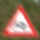} &
    \example{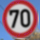} &
    \example{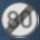}\\
    
    Target &
    \example{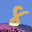} &
    \example{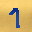} &
    \example{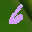} & &
    
    \example{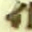} &
    \example{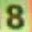} &
    \example{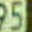} & &
    
    \example{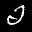} &
    \example{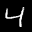} &
    \example{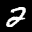} & &
    
    \example{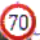} &
    \example{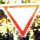} &
    \example{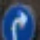}\\
    
    &
    \multicolumn{3}{c}{\rule{0pt}{0.35cm} MNIST-M} & &
    \multicolumn{3}{c}{SVHN} & &
    \multicolumn{3}{c}{MNIST} & &
    \multicolumn{3}{c}{GTSRB}\\
  \end{tabular}
  \end{small}
  \end{sc}
  \caption{Examples of domain pairs used in the experiments. See \sect{exper_quant} for details.}
  \label{fig:exper_domains_examples}
\end{figure*}

\begin{table*}[t]
  \vskip 0.15in
  \begin{center}
    \begin{small}
      \begin{sc}
        \renewcommand{\arraystretch}{1.5}
        \begin{tabular}{l r | c c c c}
          \hline
          \multirow{2}{*}{Method} & {\scriptsize Source} & MNIST & Syn Numbers & SVHN & Syn Signs \\
          & {\scriptsize Target} & MNIST-M & SVHN & MNIST & GTSRB \\
          \hline
          \multicolumn{2}{l |}{Source only} & 
          $ .5749 $                      & $ .8665 $                      & $ .5919 $                      & $ .7400 $                      \\
          \multicolumn{2}{l |}{SA \cite{Fernando13}} & 
          $ .6078 \; (7.9\%) $           & $ .8672 \; (1.3\%) $           & $ .6157 \; (5.9\%) $           & $ .7635 \; (9.1\%) $           \\
          \multicolumn{2}{l |}{Proposed approach} & 
          $ \mathbf{.8149} \; (57.9\%) $ & $ \mathbf{.9048} \; (66.1\%) $ & $ \mathbf{.7107} \; (29.3\%) $ & $ \mathbf{.8866} \; (56.7\%) $ \\
          \multicolumn{2}{l |}{Train on target} & 
          $ .9891 $                      & $ .9244 $                      & $ .9951 $                      & $ .9987 $                      \\
          \hline
        \end{tabular}
      \end{sc}
    \end{small}
  \end{center}
    \caption{Classification accuracies for digit image classifications for different source and target domains. {\sc MNIST-M} corresponds to difference-blended digits over non-uniform background. The first row corresponds to the lower performance bound (i.e.\ if no adaptation is performed). The last row corresponds to training on the target domain data with known class labels (upper bound on the DA performance). For each of the two DA methods (ours and \cite{Fernando13}) we show how much of the gap between the lower and the upper bounds was covered (in brackets). For all five cases, our approach outperforms \cite{Fernando13} considerably, and covers a big portion of the gap.\vspace{-0mm} }
  \label{tab:results}
  \vskip -0.1in
\end{table*}

\begin{table*}[t]
  \vskip 0.15in
  \begin{center}
    \begin{small}
      \begin{sc}
        \renewcommand{\arraystretch}{1.5}
        \begin{tabular}{l r | c c c}
          \hline
          \multirow{2}{*}{Method} & {\scriptsize Source} & Amazon & DSLR & Webcam \\
          & {\scriptsize Target} & Webcam & Webcam & DSLR \\
          \hline
          \multicolumn{2}{l |}{GFK(PLS, PCA) \cite{Gong12}} & 
          $ .464 \pm .005 $ & $ .613 \pm .004 $ & $ .663 \pm .004 $\\ 
          \multicolumn{2}{l |}{SA \cite{Fernando13}} & 
          $ .450 $ & $ .648 $ & $ .699 $\\ 
          \multicolumn{2}{l |}{DA-NBNN \cite{Tommasi13}} & 
          $ .528 \pm .037 $ & $ .766 \pm .017 $ & $ .762 \pm .025 $\\ 
          \multicolumn{2}{l |}{DLID \cite{Chopra13}} & 
          $ .519 $ & $ .782 $ & $ .899 $\\
          \multicolumn{2}{l |}{DeCAF$_6$ Source Only \cite{Donahue14}} &
          $ .522 \pm .017 $ & $ .915 \pm .015 $ & --\\ 
          \multicolumn{2}{l |}{DaNN \cite{Ghifary14}} & 
          $ .536 \pm .002 $ & $ .712 \pm .000 $ & $ .835 \pm .000 $\\ 
          \multicolumn{2}{l |}{DDC \cite{Tzeng14}} & 
          $ .594 \pm .008 $ & $ .925 \pm .003 $ & $ .917 \pm .008 $\\ 
          \multicolumn{2}{l |}{Proposed Approach} & 
          $ \mathbf{ .673 \pm .017 } $ & $ \mathbf{ .940 \pm .008 } $ & $ \mathbf{ .937 \pm .010 } $\\
          \hline
        \end{tabular}
      \end{sc}
    \end{small}
  \end{center}
    \caption{Accuracy evaluation of different DA approaches on the standard {\sc Office} \cite{Saenko10} dataset. Our method (last row) outperforms competitors setting the new state-of-the-art.}
  \label{tab:results_office}
\end{table*}

% Other rows refer to the following algorithms (from top to bottom): Geodesic Flow Kernel \cite{Gong12}, Subspace Alignment \cite{Fernando13}, Naive Bayes Nearest Neighbor \cite{Tommasi13},  deep learning approach from \cite{Chopra13}, DeCAF$_6$-features described in \cite{Donahue14}, Domain Adaptive NNs \cite{Ghifary14}, Deep Domain Confusion \cite{Tzeng14}.

\def\X{{\mathbf X}}
\def\y{{\mathbf y}}

% \vspace{2mm}\noindent {\bf Datasets.}
% \label{sect:exper_datasets}

% In order to test our method in the setting of traffic signs classification we obtained~100,000 synthetic images ({\sc Syn~Signs}) simulating various photoshooting conditions. This dataset was used in conjunction with {\it The German Traffic Sign Recognition Benchmark} ({\sc GTSRB}) \cite{Stallkamp12}.

% Finally, we perform domain adaption for the {\sc CIFAR-10} and the {\sc STL-10} downsampled to the size of $ 32 \times 32 $. This pair is considerably different from the previously mentioned datasets as the intra-class variability here is higher.

We perform extensive evaluation of the proposed approach on a number of popular image datasets and their modifications. These include large-scale datasets of small images popular with deep learning methods, and the {\sc Office} datasets \cite{Saenko10}, which are a {\em de facto} standard for domain adaptation in computer vision, but have much fewer images.

\vspace{2mm}\noindent {\bf Baselines.} For the bulk of experiments the following baselines are evaluated. The \textbf{source-only} model is trained without consideration for target-domain data (no domain classifier branch included into the network). The \textbf{train-on-target} model is trained on the target domain with class labels revealed. This model serves as an upper bound on DA methods, assuming that target data are abundant and the shift between the domains is considerable. 

In addition, we compare our approach against the recently proposed unsupervised DA method based on \textbf{subspace alignment (SA)} \cite{Fernando13}, which is simple to setup and test on new datasets, but has also been shown to perform very well in experimental comparisons with other ``shallow'' DA methods. To boost the performance of this baseline, we pick its most important free parameter (the number of principal components) from the range $ \{ 2, \ldots, 60 \} $, so that the test performance on the target domain is maximized. To apply SA in our setting, we train a source-only model and then consider the activations of the last hidden layer in the label predictor (before the final linear classifier) as descriptors/features, and learn the mapping between the source and the target domains \cite{Fernando13}.

Since the SA baseline requires to train a new classifier after adapting the features, and in order to put all the compared settings on an equal footing, we retrain the last layer of the label predictor using a standard linear SVM~\cite{liblinear} for all four considered methods (including ours; the performance on the target domain remains approximately the same after the retraining). 

For the {\sc Office} dataset \cite{Saenko10}, we directly compare the performance of our full network (feature extractor and label predictor) against recent DA approaches using previously published results.

\vspace{2mm}\noindent {\bf CNN architectures.} In general, we compose feature extractor from two or three convolutional layers, picking their exact configurations from previous works. We give the exact architectures in \ref{sect:appendix_archs}.

For the domain adaptator we stick to the three fully connected layers ($x\rightarrow1024\rightarrow1024\rightarrow2$), except for {\sc MNIST} where we used a simpler ($x\rightarrow100\rightarrow2$) architecture to speed up the experiments.

For loss functions, we set $ L_y $ and $ L_d $ to be the logistic regression loss and the binomial cross-entropy respectively.

\vspace{2mm}\noindent {\bf CNN training procedure.}
The model is trained on $128$-sized batches. Images are preprocessed by the mean subtraction. A half of each batch is populated by the samples from the source domain (with known labels), the rest is comprised of the target domain (with unknown labels).

In order to suppress noisy signal from the domain classifier at the early stages of the training procedure instead of fixing the adaptation factor $ \lambda $, we gradually change it from $0$ to $1$ using the following schedule:
\begin{equation}
  \lambda_p = \frac{2}{1 + \exp(-\gamma \cdot p)} - 1,
\end{equation}
where $\gamma$ was set to $10$ in all experiments (the schedule was not optimized/tweaked). Further details on the CNN training can be found in \ref{sect:appendix_training}.

\begin{figure*}
  \addtolength{\subfigcapskip}{0.1cm}
  \centering
  \begin{minipage}{.5\textwidth}
  \centering
  \small{{\sc MNIST $ \rightarrow $ MNIST-M}: top feature extractor layer}
  \setcounter{subfigure}{0}
  \hspace*{\fill}%
  \subfigure[Non-adapted]{%%
    \includegraphics[width=0.45\textwidth]{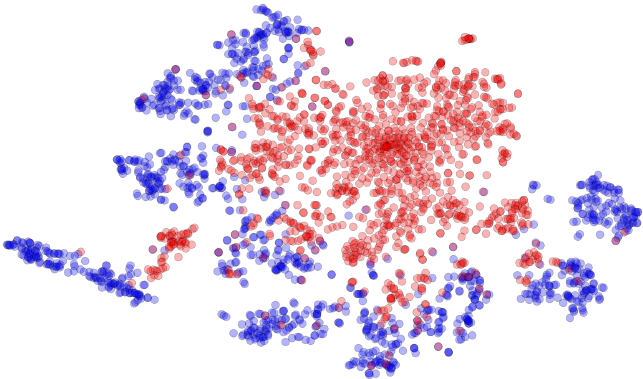}}\hfill%
  \subfigure[Adapted]{%%
    \includegraphics[width=0.45\textwidth]{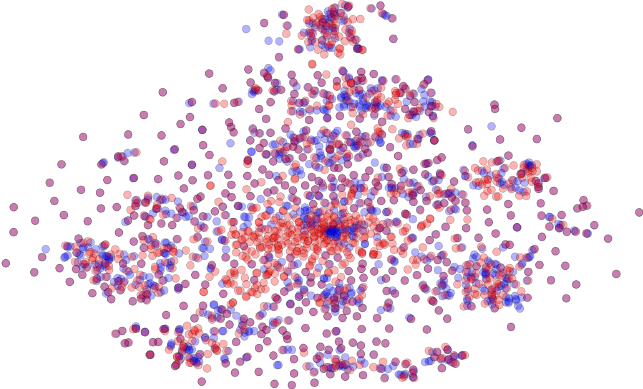}}%%
  \hspace*{\fill}%
  \end{minipage}%
  \begin{minipage}{.5\textwidth}
  \centering
  \small{{\sc Syn Numbers $ \rightarrow $ SVHN}: last hidden layer of the label predictor}
  \setcounter{subfigure}{0}
  \hspace*{\fill}%
  \subfigure[Non-adapted]{%%
    \includegraphics[width=0.45\textwidth]{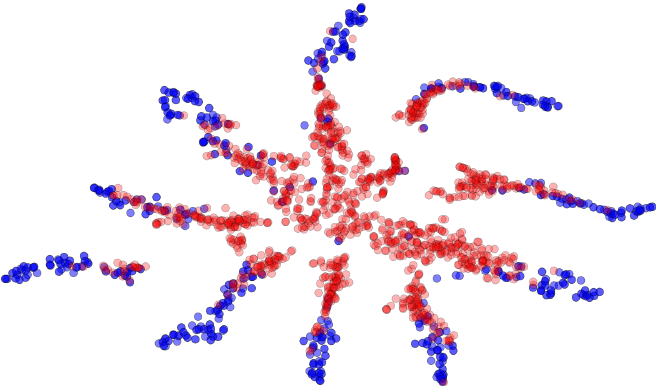}}\hfill%
  \subfigure[Adapted]{%%
    \includegraphics[width=0.45\textwidth]{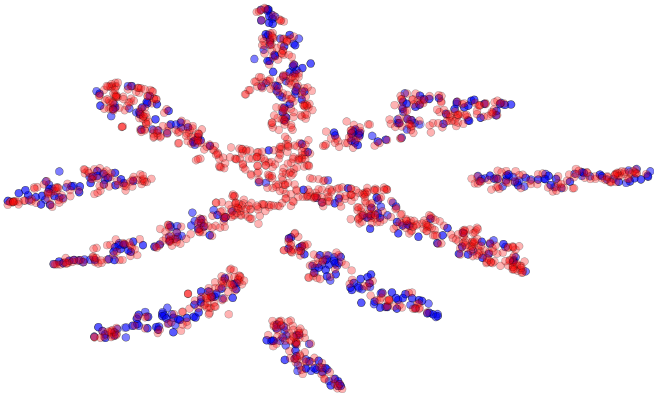}}%%
  \hspace*{\fill}%
  \end{minipage}
  \caption{The effect of adaptation on the distribution of the extracted features (best viewed in color). The figure shows t-SNE \cite{Maaten13} visualizations of the CNN's activations {\bf (a)} in case when no adaptation was performed and {\bf (b)} in case when our adaptation procedure was incorporated into training. {\it Blue} points correspond to the source domain examples, while {\it red} ones correspond to the target domain. In all cases, the adaptation in our method makes the two distributions of features much closer.}
  \label{fig:exper_adapt_vis}
\end{figure*}

\vspace{2mm}\noindent {\bf Visualizations.}
We use t-SNE \cite{Maaten13} projection to visualize feature distributions at different points of the network, while color-coding the domains (\fig{exper_adapt_vis}). We observe strong correspondence between the success of the adaptation in terms of the classification accuracy for the target domain, and the overlap between the domain distributions in such visualizations.
 
\vspace{2mm}\noindent {\bf Choosing meta-parameters.} 
In general, good unsupervised DA methods should provide ways to set meta-parameters (such as $\lambda$, the learning rate, the momentum rate, the network architecture for our method) in an unsupervised way, i.e.\ without referring to labeled data in the target domain. %Here we would like to give few recommendations concerning this matter. First, as it was pointed out in \sect{theory} the domain classifier should not be significantly more complex than the label predictor. 
In our method, one can assess the performance of the whole system (and the effect of changing hyper-parameters) by observing the test error on the source domain {\em and} the domain classifier error. In general, we observed a good correspondence between the success of adaptation and these errors (adaptation is more successful when the source domain test error is low, while the domain classifier error is high).
In addition, the layer, where the the domain adaptator is attached can be picked by computing difference between means as suggested in \cite{Tzeng14}. 

% \begin{figure*}
%   \centering
%   {\sc MNIST $ \rightarrow $ MNIST ($ | \Delta | $, bg)}: top feature extractor layer
%   \setcounter{subfigure}{0}
%   \subfigure[Non-adapted]{%%
%     \scalebox{0.8}{\input{./figures/adaptation_vis/pool2_mnist2inv_before.pgf}}}%%
%   \subfigure[Adapted]{%%
%     \scalebox{0.8}{\input{./figures/adaptation_vis/pool2_mnist2inv_after.pgf}}}\\
%   \vspace{5mm}
%   {\sc Syn Numbers $ \rightarrow $ SVHN}: last hidden layer of the label predictor
%   \setcounter{subfigure}{0}
%   \subfigure[Non-adapted]{%%
%     \scalebox{0.8}{\input{./figures/adaptation_vis/before.pgf}}}%%
%   \subfigure[Adapted]{%%
%     \scalebox{0.8}{\input{./figures/adaptation_vis/after.pgf}}}%%
%   \caption{The effect of adaptation on the distribution of the extracted features. The figure shows t-SNE \cite{Maaten13} visualizations of the CNN's activations {\bf (a)} in case when no adaptation was performed and {\bf (b)} in case when our adaptation procedure was incorporated into training. {\it Blue} points correspond to the source domain examples, while {\it red} ones correspond to the target domain. In all cases, the adaptation in our method makes the two distributions of features much closer.}
%   \label{fig:exper_adapt_vis}
% \end{figure*}

\subsection{Results}
\label{sect:exper_quant}

We now discuss the experimental settings and the results. In each case, we train on the source dataset and test on a different target domain dataset, with considerable shifts between domains (see \fig{exper_domains_examples}). The results are summarized in \tab{results} and \tab{results_office}. 

\vspace{2mm}\noindent {\bf MNIST $ \rightarrow $ MNIST-M.}
Our first experiment deals with the MNIST dataset~\cite{LeCun98} (source). In order to obtain the target domain ({\sc MNIST-M}) we blend digits from the original set over patches randomly extracted from color photos from BSDS500 \cite{Arbelaez11}. This operation is formally defined for two images $ I^{1}, I^{2} $ as $ I_{ijk}^{out} = | I_{ijk}^{1} - I_{ijk}^{2} | $, where $ i, j $ are the coordinates of a pixel and $ k $ is a channel index. In other words, an output sample is produced by taking a patch from a photo and inverting its pixels at positions corresponding to the pixels of a digit. For a human the classification task becomes only slightly harder compared to the original dataset (the digits are still clearly distinguishable) whereas for a CNN trained on MNIST this domain is quite distinct, as the background and the strokes are no longer constant. Consequently, the source-only model performs poorly. Our approach succeeded at aligning feature distributions (\fig{exper_adapt_vis}), which led to successful adaptation results (considering that the adaptation is unsupervised). At the same time, the improvement over source-only model achieved by subspace alignment (SA) \cite{Fernando13} is quite modest, thus highlighting the difficulty of the adaptation task. 

\vspace{2mm}\noindent {\bf Synthetic numbers $ \rightarrow $ SVHN.}
To address a common scenario of training on synthetic data and testing on  real data, we use Street-View House Number dataset {\sc SVHN} \cite{Netzer11} as the target domain and synthetic digits as the source. The latter ({\sc Syn ~Numbers}) consists of ~500,000 images generated by ourselves from Windows fonts by varying the text (that includes different one-, two-, and three-digit numbers), positioning, orientation, background and stroke colors, and the amount of blur. The degrees of variation were chosen manually to simulate SVHN, however the two datasets are still rather distinct, the biggest difference being the structured clutter in the background of SVHN images. 

The proposed backpropagation-based technique works well covering two thirds of the gap between training with source data only and training on target domain data with known target labels. In contrast, SA~\cite{Fernando13} does not result in any significant improvement in the classification accuracy, thus highlighting that the adaptation task is even more challenging than in the case of the MNIST experiment.

\vspace{2mm}\noindent {\bf MNIST $ \leftrightarrow $ SVHN.}
In this experiment, we further increase the gap between distributions, and test on {\sc MNIST} and {\sc SVHN}, which are significantly different in appearance. Training on SVHN even without adaptation is challenging --- classification error stays high during the first 150 epochs. In order to avoid ending up in a poor local minimum we, therefore, do not use learning rate annealing here. Obviously, the two directions ({\sc MNIST} $ \rightarrow $ {\sc SVHN} and {\sc SVHN} $ \rightarrow $ {\sc MNIST}) are not equally difficult. As {\sc SVHN} is more diverse, a model trained on SVHN is expected to be more generic and to perform reasonably on the MNIST dataset. This, indeed, turns out to be the case and is supported by the appearance of the feature distributions. We observe a quite strong separation between the domains when we feed them into the CNN trained solely on { \sc MNIST}, whereas for the {\sc SVHN}-trained network the features are much more intermixed. This difference probably explains why our method succeeded in improving the performance by adaptation in the {\sc SVHN} $ \rightarrow $ {\sc MNIST} scenario (see \tab{results}) but not in the opposite direction (SA is not able to perform adaptation in this case either). Unsupervised adaptation from MNIST to SVHN gives a failure example for our approach (we are unaware of any unsupervised DA methods capable of performing such adaptation).

\vspace{2mm}\noindent {\bf Synthetic Signs $ \rightarrow $ GTSRB.}
Overall, this setting is similar to the {\sc Syn Numbers} $ \rightarrow $ {\sc SVHN} experiment, except the distribution of the features is more complex due to the significantly larger number of classes (43 instead of 10). For the source domain we obtained~100,000 synthetic images (which we call {\sc Syn~Signs}) simulating various photoshooting conditions. Once again, our method achieves a sensible increase in performance once again proving its suitability for the synthetic-to-real data adaptation.

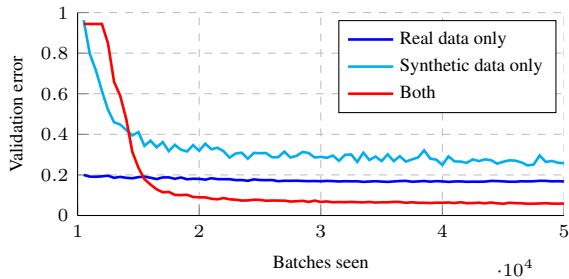
\begin{figure}
  \centering
  \setlength\figureheight{2.7cm}
  \setlength\figurewidth{6.8cm}
  % This file was created by matlab2tikz v0.5.0 running on MATLAB 8.3.
%Copyright (c) 2008--2014, Nico Schlömer <nico.schloemer@gmail.com>
%All rights reserved.
%Minimal pgfplots version: 1.3
%  http://www.mathworks.com/matlabcentral/fileexchange/22022-matlab2tikz
\begin{tikzpicture}[font=\scriptsize]

\begin{axis}[%
width=0.95092\figurewidth,
height=\figureheight,
at={(0\figurewidth,0\figureheight)},
scale only axis,
xmin=10000,
xmax=50000,
xlabel={Batches seen},
ymin=0,
ymax=1,
ylabel={Validation error},
axis x line*=bottom,
axis y line*=left,
legend style={at={($ (1,1) + (-0.1cm,-0.1cm) $)},anchor=north east,align=left,legend cell align=left,draw=black},
xmajorgrids,
ymajorgrids,
grid style={dashed}
]
\addplot [color=blue,solid,line width=1.0pt]
  table[row sep=crcr]{%
10500	0.199757996632997\\
11000	0.19162984006734\\
11500	0.190788089225589\\
12000	0.192918771043771\\
12500	0.196390993265993\\
13000	0.185527146464646\\
13500	0.190472432659933\\
14000	0.185606060606061\\
14500	0.183422769360269\\
15000	0.189051978114478\\
15500	0.191524621212121\\
16000	0.186079545454545\\
16500	0.179424452861953\\
17000	0.187684132996633\\
17500	0.187868265993266\\
18000	0.180923821548822\\
18500	0.187315867003367\\
19000	0.178661616161616\\
19500	0.18102904040404\\
20000	0.180555555555556\\
20500	0.176662457912458\\
21000	0.183791035353535\\
21500	0.179214015151515\\
22000	0.178898358585859\\
22500	0.178898358585859\\
23000	0.174479166666667\\
23500	0.174742213804714\\
24000	0.171059553872054\\
24500	0.177951388888889\\
25000	0.174794823232323\\
25500	0.174084595959596\\
26000	0.174636994949495\\
26500	0.169034090909091\\
27000	0.171191077441077\\
27500	0.170875420875421\\
28000	0.171506734006734\\
28500	0.170217803030303\\
29000	0.169244528619529\\
29500	0.169875841750842\\
30000	0.168744739057239\\
30500	0.17048085016835\\
31000	0.169454966329966\\
31500	0.167771464646465\\
32000	0.168849957912458\\
32500	0.168323863636364\\
33000	0.168718434343434\\
33500	0.165667087542088\\
34000	0.167376893939394\\
34500	0.169007786195286\\
35000	0.167140151515152\\
35500	0.165667087542088\\
36000	0.167850378787879\\
36500	0.169823232323232\\
37000	0.170691287878788\\
37500	0.16640361952862\\
38000	0.167981902356902\\
38500	0.169875841750842\\
39000	0.166771885521886\\
39500	0.169376052188552\\
40000	0.168087121212121\\
40500	0.165509259259259\\
41000	0.167718855218855\\
41500	0.168060816498317\\
42000	0.166035353535354\\
42500	0.166692971380471\\
43000	0.166429924242424\\
43500	0.167034932659933\\
44000	0.170349326599327\\
44500	0.169744318181818\\
45000	0.168218644781145\\
45500	0.166429924242424\\
46000	0.166324705387205\\
46500	0.168771043771044\\
47000	0.168034511784512\\
47500	0.168718434343434\\
48000	0.171059553872054\\
48500	0.170638678451178\\
49000	0.16819234006734\\
49500	0.168981481481481\\
50000	0.167902988215488\\
};
\addlegendentry{Real data only};

\addplot [color=cyan,solid,line width=1.0pt]
  table[row sep=crcr]{%
10500	0.9625\\
11000	0.79765625\\
11500	0.715625\\
12000	0.6140625\\
12500	0.52109375\\
13000	0.459375\\
13500	0.4484375\\
14000	0.421875\\
14500	0.39453125\\
15000	0.4109375\\
15500	0.34296875\\
16000	0.36875\\
16500	0.3359375\\
17000	0.36171875\\
17500	0.3171875\\
18000	0.3484375\\
18500	0.32421875\\
19000	0.315625\\
19500	0.346875\\
20000	0.31875\\
20500	0.35390625\\
21000	0.3265625\\
21500	0.33359375\\
22000	0.3171875\\
22500	0.28515625\\
23000	0.30546875\\
23500	0.309375\\
24000	0.2796875\\
24500	0.30859375\\
25000	0.30703125\\
25500	0.3078125\\
26000	0.28671875\\
26500	0.2875\\
27000	0.31484375\\
27500	0.2859375\\
28000	0.29375\\
28500	0.31328125\\
29000	0.3078125\\
29500	0.2859375\\
30000	0.2890625\\
30500	0.284375\\
31000	0.2953125\\
31500	0.26953125\\
32000	0.29921875\\
32500	0.30078125\\
33000	0.2640625\\
33500	0.309375\\
34000	0.2734375\\
34500	0.290625\\
35000	0.26796875\\
35500	0.3015625\\
36000	0.26796875\\
36500	0.2921875\\
37000	0.265625\\
37500	0.2765625\\
38000	0.2859375\\
38500	0.32109375\\
39000	0.28046875\\
39500	0.275\\
40000	0.24921875\\
40500	0.29140625\\
41000	0.26640625\\
41500	0.265625\\
42000	0.259375\\
42500	0.2765625\\
43000	0.26796875\\
43500	0.2765625\\
44000	0.27265625\\
44500	0.25546875\\
45000	0.26484375\\
45500	0.271875\\
46000	0.2703125\\
46500	0.26171875\\
47000	0.246875\\
47500	0.25078125\\
48000	0.29609375\\
48500	0.2640625\\
49000	0.26875\\
49500	0.26015625\\
50000	0.2578125\\
};
\addlegendentry{Synthetic data only};

\addplot [color=red,solid,line width=1.0pt]
  table[row sep=crcr]{%
10500	0.943892045454545\\
11000	0.943892045454545\\
11500	0.943892045454545\\
12000	0.943892045454545\\
12500	0.848300715488216\\
13000	0.658722643097643\\
13500	0.590593434343434\\
14000	0.475484006734007\\
14500	0.313946759259259\\
15000	0.235690235690236\\
15500	0.17879313973064\\
16000	0.152383207070707\\
16500	0.12912984006734\\
17000	0.114478114478114\\
17500	0.116214225589226\\
18000	0.1015625\\
18500	0.10066813973064\\
19000	0.101983375420875\\
19500	0.0914351851851852\\
20000	0.0895675505050505\\
20500	0.0894360269360269\\
21000	0.0827283249158249\\
21500	0.0798611111111111\\
22000	0.0859638047138047\\
22500	0.0799137205387205\\
23000	0.0778619528619529\\
23500	0.0737584175084175\\
24000	0.0742582070707071\\
24500	0.0776778198653199\\
25000	0.0771517255892256\\
25500	0.0725747053872054\\
26000	0.0739425505050505\\
26500	0.0734953703703704\\
27000	0.0730744949494949\\
27500	0.0688920454545455\\
28000	0.0702072811447811\\
28500	0.072337962962963\\
29000	0.0670244107744108\\
29500	0.0733638468013468\\
30000	0.0667613636363636\\
30500	0.0692340067340067\\
31000	0.0652093855218855\\
31500	0.0664720117845118\\
32000	0.0655776515151515\\
32500	0.0671296296296296\\
33000	0.0656039562289562\\
33500	0.0646043771043771\\
34000	0.0668665824915825\\
34500	0.0638678451178451\\
35000	0.065077861952862\\
35500	0.0649989478114478\\
36000	0.0672348484848485\\
36500	0.0668665824915825\\
37000	0.0626052188552189\\
37500	0.0652093855218855\\
38000	0.0626315235690236\\
38500	0.0627893518518518\\
39000	0.0613162878787879\\
39500	0.063236531986532\\
40000	0.0629208754208754\\
40500	0.0639467592592593\\
41000	0.0612899831649832\\
41500	0.0653409090909091\\
42000	0.0608691077441077\\
42500	0.0613425925925926\\
43000	0.0630260942760943\\
43500	0.060106271043771\\
44000	0.0638678451178451\\
44500	0.0602377946127946\\
45000	0.0577388468013468\\
45500	0.062684132996633\\
46000	0.0608164983164983\\
46500	0.0603167087542088\\
47000	0.0577651515151515\\
47500	0.0583175505050505\\
48000	0.0591329966329966\\
48500	0.0607112794612795\\
49000	0.0585805976430976\\
49500	0.0583175505050505\\
50000	0.0590540824915825\\
};
\addlegendentry{Both};

\end{axis}
\end{tikzpicture}%
  \caption{Semi-supervised domain adaptation for the traffic signs. As labeled target domain data are shown to the method, it achieves significantly lower error than the model trained on target domain data only or on source domain data only. \vspace{-4mm}}
  \label{fig:exper_semi_test}
\end{figure}

As an additional experiment, we also evaluate the proposed algorithm for semi-supervised domain adaptation, i.e.\ when one is additionally provided with a small amount of labeled target data. For that purpose we split {\sc GTSRB} into the train set (1280 random samples with labels) and the validation set (the rest of the dataset). The validation part is used solely for the evaluation and does not participate in the adaptation. The training procedure changes slightly as the label predictor is now exposed to the target data. \fig{exper_semi_test} shows the change of the validation error throughout the training. While the graph clearly suggests that our method can be used in the semi-supervised setting, thorough verification of semi-supervised setting is left for future work.

\vspace{2mm}\noindent {\bf Office dataset.} 
We finally evaluate our method on {\sc Office} dataset, which is a collection of three distinct domains: {\sc Amazon}, {\sc DSLR}, and {\sc Webcam}. Unlike previously discussed datasets, {\sc Office} is rather small-scale with only 2817 labeled images spread across 31 different categories in the largest domain. The amount of available data is crucial for a successful training of a deep model, hence we opted for the fine-tuning of the CNN pre-trained on the ImageNet \cite{Jia14} as it is done in some recent DA works \cite{Donahue14,Tzeng14,Hoffman14}. We make our approach more comparable with \cite{Tzeng14} by using exactly the same network architecture replacing domain mean-based regularization with the domain classifier.

Following most previous works, we evaluate our method using 5 random splits for each of the 3 transfer tasks commonly used for evaluation. Our training protocol is close to \cite{Tzeng14,Saenko10,Gong12} as we use the same number of labeled source-domain images per category. Unlike those works and similarly to e.g.\ DLID~\cite{Chopra13} we use the whole unlabeled target domain (as the premise of our method is the abundance of unlabeled data in the target domain). Under this transductive setting, our method is able to improve previously-reported state-of-the-art accuracy for unsupervised adaptation very considerably (\tab{results_office}), especially in the most challenging {\sc Amazon} $ \rightarrow $ {\sc Webcam} scenario (the two domains with the largest domain shift).

\section{Discussion}

We have proposed a new approach to unsupervised domain adaptation of deep feed-forward architectures, which allows large-scale training based on large amount of annotated data in the source domain and large amount of unannotated data in the target domain. Similarly to many previous shallow and deep DA techniques, the adaptation is achieved through aligning the distributions of features across the two domains. However, unlike previous approaches, the alignment is accomplished through standard backpropagation training. The approach is therefore rather scalable, and can be implemented using any deep learning package. To this end we plan to release the source code for the Gradient Reversal layer along with the usage examples as an extension to \texttt{Caffe}~\cite{Jia14}.

Further evaluation on larger-scale tasks and in semi-supervised settings constitutes future work. It is also interesting whether the approach can benefit from a good initialization of the feature extractor. For this, a natural choice would be to use deep autoencoder/deconvolution network trained on both domains (or on the target domain) in the same vein as \cite{Glorot11,Chopra13}, effectively using \cite{Glorot11,Chopra13} as an initialization to our method.\newpage

\appendix
\renewcommand\thesection{Appendix \Alph{section}}
\def\x{{\mathbf x}}
\def\f{{\mathbf f}}

\def\S{{\cal S}}
\def\T{{\cal T}}

\def\R{{\mathds R}}

\def\tf{{\theta_f}}
\def\td{{\theta_d}}
\def\ty{{\theta_y}}
\def\htf{{\hat\theta_f}}
\def\htd{{\hat\theta_d}}
\def\hty{{\hat\theta_y}}

\section{An alternative optimization approach}
\label{sect:appendix_alternative}

There exists an alternative construction (inspired by \cite{Goodfellow14}) that leads to the same updates \eq{upd1}-\eq{upd3}. Rather than using the gradient reversal layer, the construction introduces two different loss functions for the domain classifier. Minimization of the first domain loss ($ L_{d+} $) should lead to a better domain discrimination, while the second domain loss ($ L_{d-} $) is minimized when the domains are distinct. Stochastic updates for $ \theta_f $ and $ \theta_d $ are then defined as:
\begin{align*}
  \tf \quad &\longleftarrow \quad \tf \;-\; \mu \left(\frac{\partial L^i_y}{\partial \tf} + \frac{\partial L^i_{d-}}{\partial \tf} \right)\\
  \td \quad &\longleftarrow \qquad \td \;-\; \mu \frac{\partial L^i_{d+}}{\partial \td} \, ,
\end{align*}
Thus, different parameters participate in the optimization of different losses

In this framework, the gradient reversal layer constitutes a special case, corresponding  to the pair of domain losses $ (L_d, -\lambda L_d) $. However, other pairs of loss functions can be used. One example would be the binomial cross-entropy \cite{Goodfellow14}:
\begin{equation*}
  L_{d+}(q, d) = \sum_{i = 1..N} d_i \log(q_i) + (1 - d_i) \log(1 - q_i) \, ,
\end{equation*}
where $ d $ indicates domain indices and $ q $ is an output of the predictor. In that case ``adversarial'' loss is easily obtained by swapping domain labels, i.e.\ $ L_{d-}(q, d) = L_{d+}(q, 1 - d) $. This particular pair has a potential advantage of producing stronger gradients at early learning stages if the domains are quite dissimilar. In our experiments, however, we did not observe any significant improvement resulting from this choice of losses.

\section{CNN architectures}
\label{sect:appendix_archs}

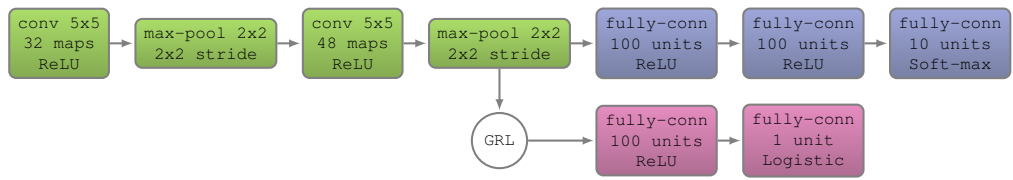
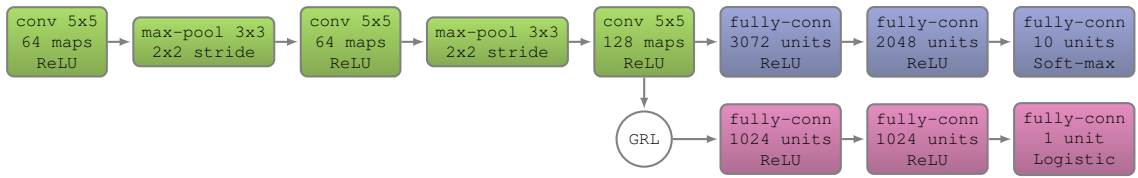
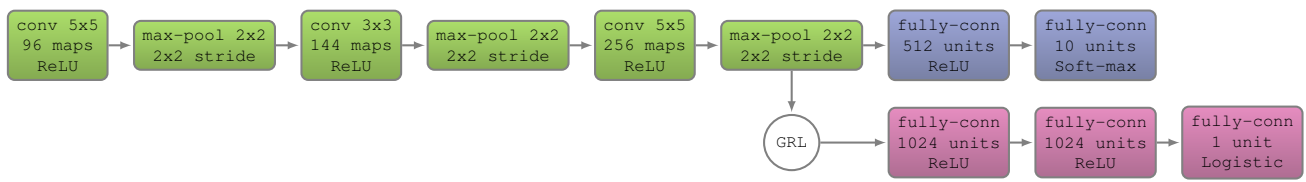
\begin{figure*}[t]
  \definecolor{fnodebottom}{RGB}{132,170,81}
  \definecolor{fnodetop}{RGB}{172,222,106}
  \definecolor{cnodebottom}{RGB}{120,128,164}
  \definecolor{cnodetop}{RGB}{158,167,218}
  \definecolor{dnodebottom}{RGB}{174,109,146}
  \definecolor{dnodetop}{RGB}{230,141,192}
  \centering
  \subfigure[MNIST architecture]{%
    \scalebox{0.65}{\begin{tikzpicture}[
  black!50, text=black,
  node distance=4mm,
  grlnode/.style={
    align=center,
    circle,minimum size=6mm,
    inner sep=5pt,
    very thick,draw=black!50,
    font=\ttfamily
  },
  fnode/.style={
    align=center,
    % The shape:
    rectangle,minimum size=6mm,rounded corners,
    % The rest
    inner sep=5pt,
    very thick,draw=black!50,
    top color=fnodetop,bottom color=fnodebottom,
    font=\ttfamily},
  cnode/.style={
    fnode,top color=cnodetop,bottom color=cnodebottom},
  dnode/.style={
    fnode,top color=dnodetop,bottom color=dnodebottom},
  vhedge/.style={
    rounded corners,to path=|- (\tikztotarget)}]
  \matrix[row sep=5mm,column sep=5mm] {
    \node (conv1) [fnode] {conv 5x5\\32 maps\\ReLU}; &
    \node (pool1) [fnode] {max-pool 2x2\\2x2 stride}; &
    \node (conv2) [fnode] {conv 5x5\\48 maps\\ReLU}; &
    \node (pool2) [fnode] {max-pool 2x2\\2x2 stride}; &
  
    \node (fc3)   [cnode] {fully-conn\\100 units\\ReLU}; &
    \node (fc4)   [cnode] {fully-conn\\100 units\\ReLU}; &
    \node (fc5)   [cnode] {fully-conn\\10 units\\Soft-max}; \\
  
    & & &
    \node (grl) [grlnode] {GRL}; &
  
    \node (fc1_d) [dnode] {fully-conn\\100 units\\ReLU}; &
    \node (fc2_d) [dnode] {fully-conn\\1 unit\\Logistic}; \\
  };
  
  \path (conv1) edge[-latex,shorten >=1pt,very thick] (pool1);
  \path (pool1) edge[-latex,shorten >=1pt,very thick] (conv2);
  \path (conv2) edge[-latex,shorten >=1pt,very thick] (pool2);
  \path (pool2) edge[-latex,shorten >=1pt,very thick] (fc3);
  \path (fc3)   edge[-latex,shorten >=1pt,very thick] (fc4);
  \path (fc4)   edge[-latex,shorten >=1pt,very thick] (fc5);
  
  \path (pool2.south) edge[-latex,shorten >=1pt,very thick] (grl.north);
  \path (grl) edge[-latex,shorten >=1pt,very thick] (fc1_d);
  \path (fc1_d) edge[-latex,shorten >=1pt,very thick] (fc2_d);
\end{tikzpicture}}}\\
  \subfigure[SVHN architecture]{%
    \scalebox{0.65}{\begin{tikzpicture}[
  black!50, text=black,
  node distance=4mm,
  grlnode/.style={
    align=center,
    circle,minimum size=6mm,
    inner sep=5pt,
    very thick,draw=black!50,
    font=\ttfamily
  },
  fnode/.style={
    align=center,
    % The shape:
    rectangle,minimum size=6mm,rounded corners,
    % The rest
    inner sep=5pt,
    very thick,draw=black!50,
    top color=fnodetop,bottom color=fnodebottom,
    font=\ttfamily},
  cnode/.style={
    fnode,top color=cnodetop,bottom color=cnodebottom},
  dnode/.style={
    fnode,top color=dnodetop,bottom color=dnodebottom},
  vhedge/.style={
    rounded corners,to path=|- (\tikztotarget)}]
  \matrix[row sep=5mm,column sep=5mm] {
    \node (conv1) [fnode] {conv 5x5\\64 maps\\ReLU}; &
    \node (pool1) [fnode] {max-pool 3x3\\2x2 stride}; &
    \node (conv2) [fnode] {conv 5x5\\64 maps\\ReLU}; &
    \node (pool2) [fnode] {max-pool 3x3\\2x2 stride}; &
    \node (conv3) [fnode] {conv 5x5\\128 maps\\ReLU}; &
  
    \node (fc4)   [cnode] {fully-conn\\3072 units\\ReLU}; &
    \node (fc5)   [cnode] {fully-conn\\2048 units\\ReLU}; &
    \node (fc6)   [cnode] {fully-conn\\10 units\\Soft-max}; \\
  
    & & & &
    \node (grl) [grlnode] {GRL}; &
    \node (fc1_d) [dnode] {fully-conn\\1024 units\\ReLU}; &
    \node (fc2_d) [dnode] {fully-conn\\1024 units\\ReLU}; &
    \node (fc3_d) [dnode] {fully-conn\\1 unit\\Logistic}; \\
  };
  
  \path (conv1) edge[-latex,shorten >=1pt,very thick] (pool1);
  \path (pool1) edge[-latex,shorten >=1pt,very thick] (conv2);
  \path (conv2) edge[-latex,shorten >=1pt,very thick] (pool2);
  \path (pool2) edge[-latex,shorten >=1pt,very thick] (conv3);
  \path (conv3) edge[-latex,shorten >=1pt,very thick] (fc4);
  \path (fc4)   edge[-latex,shorten >=1pt,very thick] (fc5);
  \path (fc5)   edge[-latex,shorten >=1pt,very thick] (fc6);
  
  \path (conv3.south) edge[-latex,shorten >=1pt,very thick] (grl.north);
  \path (grl) edge[-latex,shorten >=1pt,very thick] (fc1_d);
  \path (fc1_d) edge[-latex,shorten >=1pt,very thick] (fc2_d);
  \path (fc2_d) edge[-latex,shorten >=1pt,very thick] (fc3_d);
\end{tikzpicture}}}\\
  \subfigure[GTSRB architecture]{%
    \scalebox{0.65}{\begin{tikzpicture}[
  black!50, text=black,
  node distance=4mm,
  grlnode/.style={
    align=center,
    circle,minimum size=6mm,
    inner sep=5pt,
    very thick,draw=black!50,
    font=\ttfamily
  },
  fnode/.style={
    align=center,
    % The shape:
    rectangle,minimum size=6mm,rounded corners,
    % The rest
    inner sep=5pt,
    very thick,draw=black!50,
    top color=fnodetop,bottom color=fnodebottom,
    font=\ttfamily},
  cnode/.style={
    fnode,top color=cnodetop,bottom color=cnodebottom},
  dnode/.style={
    fnode,top color=dnodetop,bottom color=dnodebottom},
  vhedge/.style={
    rounded corners,to path=|- (\tikztotarget)}]
  \matrix[row sep=5mm,column sep=5mm] {
    \node (conv1) [fnode] {conv 5x5\\96 maps\\ReLU}; &
    \node (pool1) [fnode] {max-pool 2x2\\2x2 stride}; &
    \node (conv2) [fnode] {conv 3x3\\144 maps\\ReLU}; &
    \node (pool2) [fnode] {max-pool 2x2\\2x2 stride}; &
    \node (conv3) [fnode] {conv 5x5\\256 maps\\ReLU}; &
    \node (pool3) [fnode] {max-pool 2x2\\2x2 stride}; &
  
    \node (fc4)   [cnode] {fully-conn\\512 units\\ReLU}; &
    \node (fc5)   [cnode] {fully-conn\\10 units\\Soft-max}; \\
  
    & & & & &
    \node (grl) [grlnode] {GRL}; &
    \node (fc1_d) [dnode] {fully-conn\\1024 units\\ReLU}; &
    \node (fc2_d) [dnode] {fully-conn\\1024 units\\ReLU}; &
    \node (fc3_d) [dnode] {fully-conn\\1 unit\\Logistic}; \\
  };
  
  \path (conv1) edge[-latex,shorten >=1pt,very thick] (pool1);
  \path (pool1) edge[-latex,shorten >=1pt,very thick] (conv2);
  \path (conv2) edge[-latex,shorten >=1pt,very thick] (pool2);
  \path (pool2) edge[-latex,shorten >=1pt,very thick] (conv3);
  \path (conv3) edge[-latex,shorten >=1pt,very thick] (pool3);
  \path (pool3) edge[-latex,shorten >=1pt,very thick] (fc4);
  \path (fc4)   edge[-latex,shorten >=1pt,very thick] (fc5);
  
  \path (pool3.south) edge[-latex,shorten >=1pt,very thick] (grl.north);
  \path (grl) edge[-latex,shorten >=1pt,very thick] (fc1_d);
  \path (fc1_d) edge[-latex,shorten >=1pt,very thick] (fc2_d);
  \path (fc2_d) edge[-latex,shorten >=1pt,very thick] (fc3_d);
\end{tikzpicture}}}\\
%   \subfigure[CIFAR-10 architecture]{%
%     \scalebox{0.65}{\input{./figures/archs/cifar10.tikz}}}
  \caption{CNN architectures used in the experiments. Boxes correspond to transformations applied to the data. Color-coding is the same as in \fig{arch}.}
  \label{fig:exper_archs}
\end{figure*}

Four different architectures were used in our experiments (first three are shown in \fig{exper_archs}):
\begin{itemize}
  \item A smaller one (a) if the source domain is MNIST. This architecture was inspired by the classical LeNet-5 \cite{LeCun98}.
  \item (b) for the experiments involving SVHN dataset. This one is adopted from \cite{Srivastava14}.
  \item (c) in the {\sc Syn Sings} $ \rightarrow $ {\sc GTSRB} setting. We used the single-CNN baseline from \cite{Cirecsan12} as our starting point.
  \item Finally, we use pre-trained \texttt{AlexNet} from the \texttt{Caffe}-package \cite{Jia14} for the {\sc Office} domains. Adaptation architecture is identical to \cite{Tzeng14}: 2-layer domain classifier ($x\rightarrow1024\rightarrow1024\rightarrow2$) is attached to the $ 256 $-dimensional bottleneck of \texttt{fc7}.  
\end{itemize}
The domain classifier branch in all cases is somewhat arbitrary (better adaptation performance might be attained if this part of the architecture is tuned).

\section{Training procedure}
\label{sect:appendix_training}

We use stochastic gradient descent with 0.9 momentum and the learning rate annealing described by the following formula:
\begin{equation*}
  \mu_p = \frac{\mu_0}{(1 + \alpha \cdot p)^\beta} \, , 
\end{equation*}
where $ p $ is the training progress linearly changing from 0 to 1, $ \mu_0 = 0.01 $, $ \alpha = 10 $ and $ \beta = 0.75 $ (the schedule was optimized to promote convergence and low error on the \emph{source} domain).

Following \cite{Srivastava14} we also use dropout and $ \ell_2 $-norm restriction when we train the SVHN architecture.

\newpage
\bibliography{paper}
\bibliographystyle{icml2015}

\end{document}